\documentclass[10pt,twocolumn,letterpaper]{article}

\usepackage{cvpr}              %

\usepackage{graphicx}
\usepackage{amsmath}
\usepackage{amssymb}
\usepackage{booktabs}
\usepackage{multirow}

\usepackage{diagbox}

\usepackage{subcaption}

\def \paravspace {-1\baselineskip}

\usepackage{array}
\newcolumntype{x}[1]{>{\centering\arraybackslash\hspace{0pt}}m{#1}}

\usepackage[pagebackref,breaklinks,colorlinks]{hyperref}

\usepackage[capitalize]{cleveref}
\crefname{section}{Sec.}{Secs.}
\Crefname{section}{Section}{Sections}
\Crefname{table}{Table}{Tables}
\crefname{table}{Tab.}{Tabs.}

\begin{document}

\title{Audiovisual Masked Autoencoders}

\author{
	Mariana-Iuliana Georgescu\textsuperscript{1, 2}\thanks{Equal contribution. Correspondence to aarnab@google.com.}  \thanks{Work done during an internship at Google.}	\qquad Eduardo Fonseca\textsuperscript{1}	\qquad Radu Tudor Ionescu\textsuperscript{2}	\\
    Mario Lucic\textsuperscript{1} 
	\qquad Cordelia Schmid\textsuperscript{1}	\qquad Anurag Arnab\textsuperscript{1}\footnotemark[1] \\
	\textsuperscript{1}Google Research \qquad \textsuperscript{2}University of Bucharest \\
}

\maketitle

\begin{abstract}

Can we leverage the audiovisual information already present in video to improve self-supervised representation learning?
To answer this question, we study various pretraining architectures and objectives within the masked autoencoding framework, motivated by the success of similar methods in natural language and image understanding.
We show that we can achieve significant improvements on audiovisual downstream classification tasks, surpassing the state-of-the-art on VGGSound and AudioSet.
Furthermore, we can leverage our audiovisual pretraining scheme for multiple unimodal downstream tasks using a single audiovisual pretrained model.
We additionally demonstrate the transferability of our representations, achieving state-of-the-art audiovisual results on Epic Kitchens without pretraining specifically for this dataset.
To facilitate further research, we have released code and models at \href{https://github.com/google-research/scenic/tree/main/scenic/projects/av_mae}{https://github.com/google-research/scenic}.
\vspace{-0.5\baselineskip}

\end{abstract}

\vspace{-0.5\baselineskip}
\section{Introduction}
\vspace{-1mm}

\begin{figure}[t]
    \centering
    \vspace{-1.5\baselineskip}
    \includegraphics[width=0.8\linewidth]{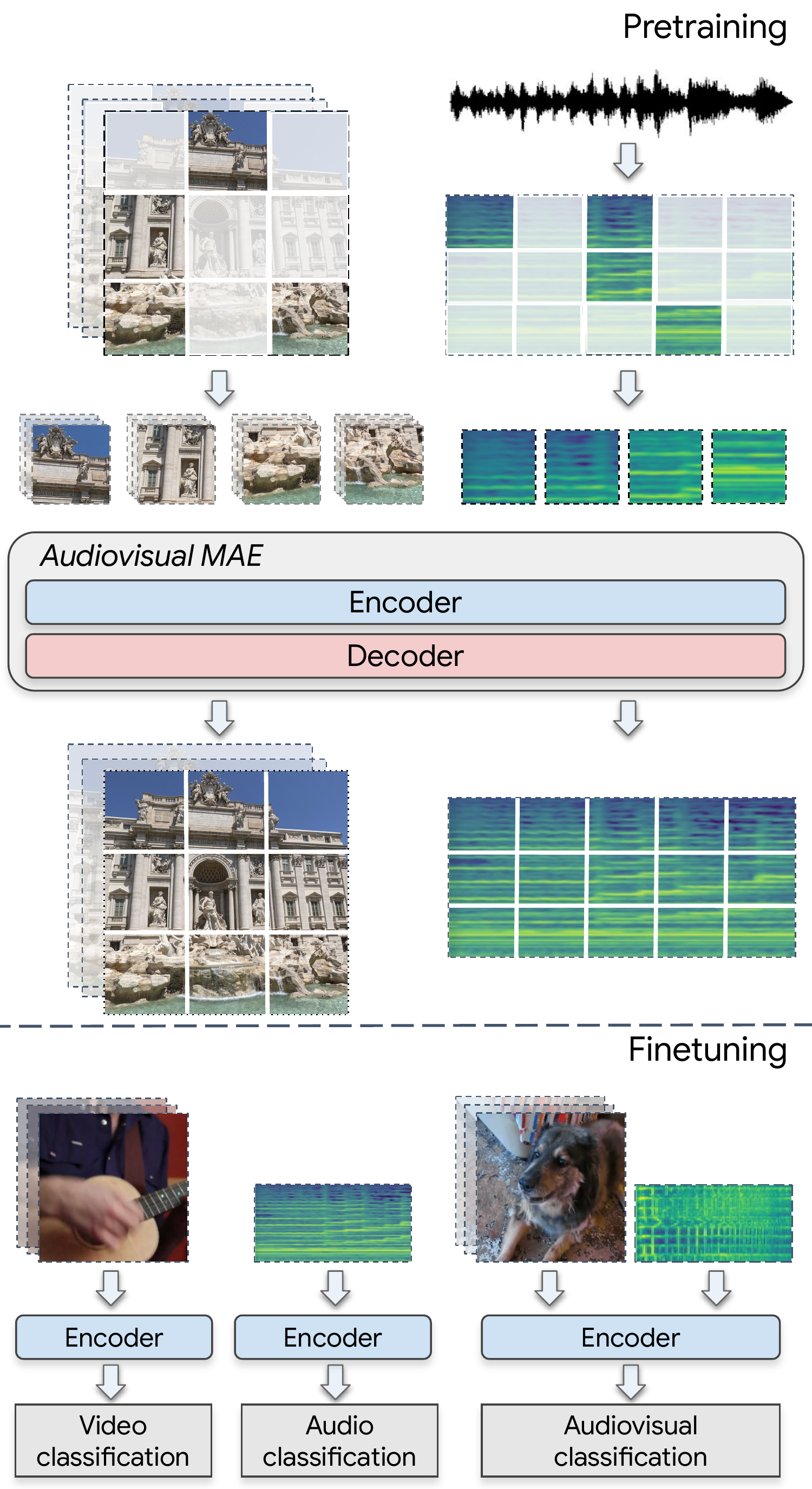}
    \vspace{-0.5\baselineskip}
    \caption{
    	Overview of our Audiovisual Masked Autoencoder.
    	We jointly encode and reconstruct audiovisual inputs, to leverage the correlations between the two modalities to learn stronger representations of the data.
    	Our pretrained encoder can then be used for audiovisual, audio-only and video-only downstream tasks.
        \vspace{-2\baselineskip}
	}
    \label{fig:teaser}
\end{figure}

The computer vision community has witnessed rapid progress across a wide range of tasks and domains, driven by progressively larger models and datasets~\cite{dosovitskiy_iclr_2021,radford2021learning,bommasani2021opportunities,yuan2021florence,singh2022flava}.
However, pretraining models on large labelled datasets~\cite{deng_cvpr_2009,sun_iccv_2017,mahajan_eccv_2018} and then finetuning on smaller target datasets is not scalable: %
Annotating large pretraining datasets is expensive and time-consuming, and larger, more performant models require more pretraining data~\cite{dosovitskiy_iclr_2021}.
This has led to growing research in self-supervised pretraining methods which learn feature representations from unlabelled data, and has been extremely successful in natural language processing (NLP) for developing large language models~\cite{brown2020language, devlin_naacl_2019, smith2022using}.
More recently, similar methods have been adopted in the vision community as well~\cite{bao2021beit,he2022masked,wei2022masked}.

In this paper, we propose to leverage the audiovisual information present in video to improve self-supervised representation learning. %
Despite recent advances in self-supervised image-~\cite{he2022masked, bao2021beit} and video-representation learning~\cite{feichtenhofer2022masked,wei2022masked,tong2022videomae}, these works still ignore the additional auditory information that is already present in their pretraining sets. %
Intuitively, we aim to exploit the correlations between the modalities already present in video to learn stronger representations of the data for unimodal and multimodal downstream tasks.
Our approach is further motivated by the fact that the world, and human perception of it, is inherently multimodal~\cite{smith2005development, shams2010crossmodal}.%

Our approach is inspired by the masked autoencoding framework~\cite{he2022masked, bao2021beit}, which itself is based on similar masked data modelling approaches in NLP~\cite{devlin_naacl_2019} and earlier works on denoising autoencoders~\cite{vincent2008extracting,pathak2016context}.
We develop multiple pretraining architectures to jointly encode and reconstruct audiovisual inputs, and conduct thorough ablation studies to verify our design choices.
To encourage further cross-modal information modelling, we also propose a novel ``inpainting'' objective which tasks our transformer model with predicting audio from video tokens and vice versa.

Our audiovisual pretraining enables us to achieve state-of-the-art results in downstream, audiovisual datasets such as VGGSound and AudioSet.
Moreover, we show how we can reuse our audiovisual pretraining for unimodal, \ie audio-only or video-only downstream classification tasks.
Furthermore, we show how the representations learned by our model transfer between different pretraining and downstream finetuning datasets, enabling us to achieve state-of-the-art art results on the Epic Kitchens dataset.

\vspace{-0.45\baselineskip}
\section{Related Work}
\vspace{-0.35\baselineskip}

Early self-supervised learning works in vision were based on solving hand-designed pretext tasks such as relative patch prediction~\cite{doersch2015unsupervised} and colourisation~\cite{zhang2016colorful}. %
Subsequent works converged on contrastive~\cite{chen2020asimple, hadsell2006dimensionality, he2020momentum, oord2018representation}, self-distillation~\cite{grill2020bootstrap, caron2021emerging, lee2021compressive} or clustering-based~\cite{caron2018deep, asano2020self} objectives that encouraged a neural network to learn feature representations that were invariant to a predefined set of data transformations.
These ideas were extended to multimodal scenarios as well, by predicting whether visual and audio signals come from the same video~\cite{arandjelovic2017look, arandjelovic2018objects,owens2018audio,korbar2018cooperative}, by audiovisual clustering~\cite{alwassel2020self,asano2020labelling} or by using contrastive losses to encourage different modalities from the same input to have similar learned embeddings~\cite{akbari2021vatt, alayrac2020self, miech2020end, patrick2021compositions, zellers2021merlot, wang2021multimodal}.

Our approach, in contrast, is based on masked data modelling -- a paradigm which removes part of the input data and then learns to predict this removed content -- and has gained traction due to the success of BERT~\cite{devlin_naacl_2019} in NLP.
BEIT~\cite{bao2021beit} notably adopted BERT-style pretraining for images, using a discrete variational autoencoder~\cite{ramesh2021zero} to produce a vocabulary of image tokens, and then predicting these tokens for masked image patches using a cross-entropy loss.
Masked Autoencoders (MAE)~\cite{he2022masked} further showed that simply regressing to the original inputs in pixel-space was just as effective, and  by only processing unmasked tokens in the encoder, training could be significantly accelerated.
MAE has recently been extended to video~\cite{feichtenhofer2022masked,tong2022videomae,wang2022bevt} and audio~\cite{chong2022masked,xu2022masked}.
Our work also uses the masked autoencoding framework, but jointly models both audio and video, and is demonstrated on both unimodal (\ie video-only and audio-only) and audiovisual downstream tasks where it outperforms supervised pretraining. 

A few recent works have also addressed multimodal pretraining with MAE:
OmniMAE~\cite{girdhar2022omnimae} trains a single MAE model to reconstruct both images and video with shared weights among the modalities.
The model is trained in a multi-task setting, where it can process either images or videos, but only one modality at a time.
It is thus a self-supervised equivalent of multi-task models like~\cite{likhosherstov2021polyvit, girdhar2022omnivore} which can perform multiple classification tasks, but only whilst processing a single task from a single modality at a time.
Our model, in contrast, is developed to fuse information from different modalities, for both multimodal and unimodal downstream tasks.
Bachmann~\etal~\cite{bachmann2022multimae} develop an MAE model for dense prediction tasks, where the model reconstructs images, depth maps and segmentation maps of the image.
Training this model, however, requires real or pseudo-labels for segmentation and depth, and hence, the model is not purely self-supervised like our approach.

We also note that concurrent works, CAV-MAE~\cite{gong2022contrastive} and MAViL,~\cite{huang2022mavil}, have also recently explored joint masked autoencoding of audio and video.
We observe two main differences with respect to our current study.
Here, we focus on a thorough comparison of multiple architectures and objectives for audiovisual masked autoencoders.
In contrast, those works explore more complex systems featuring additional objectives and training methodologies such as contrastive learning~\cite{gong2022contrastive, huang2022mavil} and/or additional distillation~\cite{huang2022mavil} training stages, whilst we perform a single-stage of self-supervised pretraining.
Second, in contrast to~\cite{gong2022contrastive, huang2022mavil}, our representations are learned from random initialisation and not from ImageNet-pretrained checkpoints.

Finally, we note that vision transformer architectures~\cite{vaswani_neurips_2017, dosovitskiy_iclr_2021} have also been extended to fusing multiple modalities~\cite{nagrani2021attention,jaegle2021perceiver, ramazanova2022owl}.
Our work focuses on pretraining audiovisual models, and can in fact be used to initialise MBT~\cite{nagrani2021attention} to improve the results that the original authors achieved with supervised pretraining.

\vspace{-0.2\baselineskip}
\section{Audiovisual Masked Autoencoders}
\label{sec:method}
\vspace{-0.2\baselineskip}

\begin{figure*}[t]
    \centering
    \vspace{-0.7\baselineskip}
    \includegraphics[width=0.99\linewidth]{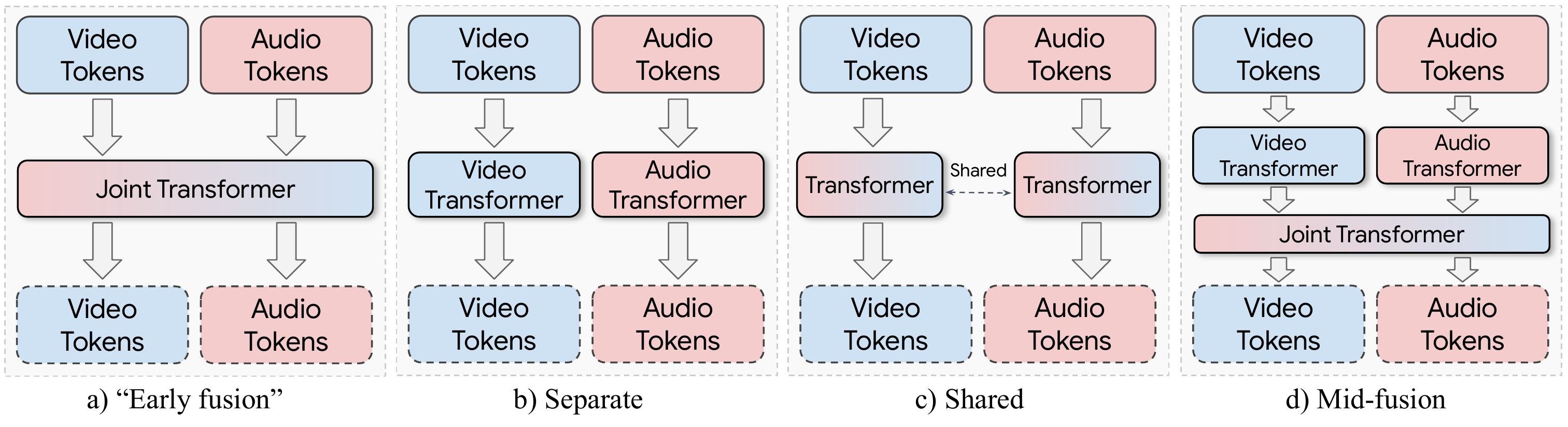} %
    \vspace{-0.7\baselineskip}
    \caption{
    	Transformer architectures for performing audiovisual fusion.
    	Concatenating the tokens before passing them through the transformer corresponds  to ``early fusion'' (a), whilst using two separate encoders (b) can be used to perform ``late fusion'' in the subsequent decoder.
    	An alternate method of coupling modalities together is by sharing weights between the two encoders (c).
    	Finally, mid-fusion (d) represents a balance between ``early'' and ``late'' fusion.
	}
	\vspace{-1.5\baselineskip}
    \label{fig:model_diagrams}
\end{figure*}

\begin{figure*}[t]
    \centering
    \vspace{-1\baselineskip}
    \includegraphics[width=0.99\linewidth]{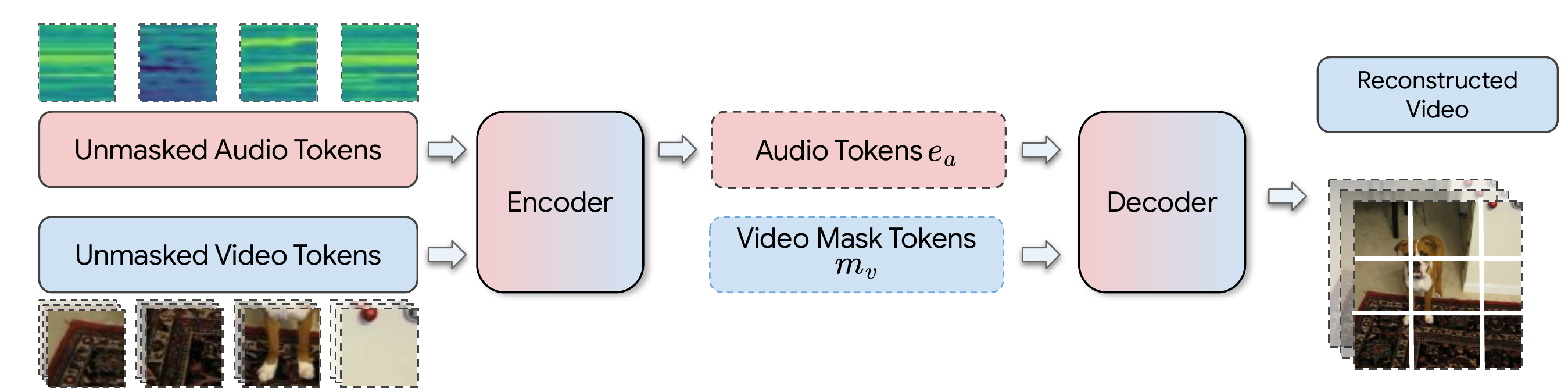}
    \vspace{-0.6\baselineskip}
    \caption{
    	Overview of modality inpainting for reconstructing video from audio.
    	We initially jointly encode unmasked tokens from both audio and video.
    	Then, we use all the encoded tokens of one modality (\ie audio), and mask tokens from the other (\ie video), to reconstruct the masked modality (\ie video).
    	Note that we can reconstruct all combinations of modalities, and show one for clarity.
    		}
	\vspace{-1.5\baselineskip}
    \label{fig:model_inpainting}
\end{figure*}

We extend the masked-autoencoding framework~\cite{he2022masked} to learn audiovisual feature representations that can be leveraged for both multimodal and unimodal downstream tasks.
This is done by jointly modelling both modalities to learn synergies between them.
We begin with an overview of masked autoencoders~\cite{he2022masked} and transformers in vision in Sec.~\ref{sec:method_background}.
Thereafter, we detail our different encoders (Sec.~\ref{sec:method_encoders}), decoders (Sec.~\ref{sec:method_decoders}) and objective functions (Sec.~\ref{sec:method_objective}) as summarised in Fig.~\ref{fig:model_diagrams} and~\ref{fig:model_inpainting}.

\vspace{-0.2\baselineskip}
\subsection{Background}
\label{sec:method_background}
\vspace{-0.2\baselineskip}

Transformers are a generic architecture that operate on any input that can be converted into a sequence of tokens.
Images are typically tokenised by performing a linear projection of non-overlapping ``patches'' of the input, which corresponds to a 2D convolution~\cite{dosovitskiy_iclr_2021}.
For videos, a common method is to linearly project spatio-temporal ``tubelets'' which is equivalent to a 3D convolution~\cite{arnab2021vivit}.
Audio inputs are commonly represented as spectrograms, which are 2-dimensional representations (along the time and frequency axes) %
in the Fourier domain of the input waveform, and can thus be treated as images with a single channel~\cite{gong2021ast} (in fact, ImageNet pretrained weights have also been shown to be effective for initialising audio models~\cite{gong2021ast,gwardys2014deep,guzhov2021esresnet}). 

Dosovitskiy~\etal~\cite{dosovitskiy_iclr_2021} showed that transformers, using the same original architecture as~\cite{vaswani_neurips_2017}, excelled at image recognition tasks when pretrained on large, supervised datasets such as ImageNet-21K~\cite{deng_cvpr_2009} and JFT~\cite{sun_iccv_2017}.
More recently, approaches such as masked autoencoders~\cite{he2022masked} have demonstrated how vision transformers can be pretrained with only self-supervision on smaller datasets.

In the masked autoencoding framework~\cite{he2022masked}, the input, $\mathbf{x}$, is tokenised following previous supervised learning setups~\cite{dosovitskiy_iclr_2021,arnab2021vivit,gong2021ast}.
We denote the resulting tokens as $\mathbf{v} = \text{Tokenise}(\mathbf{x}) + \mathbf{p},$ where $\mathbf{p}$ denotes the positional embeddings, and
$\mathbf{v} \in \mathbb{R}^{n \times d}$ where $n$ is the total number of tokens, and $d$ is their hidden dimensionality.
A random subset of these tokens are then masked, and only the unmasked tokens are processed by the transformer encoder.
We denote these steps as $\mathbf{u} = \text{Mask}(\mathbf{v} ; \alpha)$ and $\mathbf{e} = \text{Encode}(\mathbf{u})$, where $\mathbf{u}$ and $\mathbf{e} \in \mathbb{R}^{u \times d}$.
Here $\alpha$ is the masking ratio, and $u = (1 - \alpha) \cdot n$ is the number of unmasked tokens.

Learned mask tokens, $m \in \mathbb{R}^{d}$ are then inserted back into the input token sequence whilst also adding new positional embeddings, which we denote as $\mathbf{z} = \text{Unshuffle}(\mathbf{e}, m) \in \mathbb{R}^{n \times d}$. 
Finally, a transformer decoder (which has the same structure as the encoder) processes these tokens, and the entire network is trained to reconstruct the original inputs corresponding to the tokens in pixel space, $\mathbf{\tilde{x}}$, with a mean-squared error objective.
Note that $\mathbf{\tilde{x}}$ denotes the ``patchified'' version of the input $\mathbf{x}$ used as the reconstruction target.
For example, for an image $\mathbf{x} \in \mathbb{R}^{H \times W \times 3}$,  $\mathbf{\tilde{x}} \in \mathbb{R}^{ H / p_h \cdot W / p_w \times  p_h \cdot p_w \cdot 3}$, where $H$ and $W$ are the height and width of the image, and $p_h$ and $p_w$ are the patch sizes used for tokenisation.
Additional standardisation may also be applied to $\mathbf{\tilde{x}}$~\cite{he2022masked,tong2022videomae}.

\subsection{Audiovisual encoders}
\label{sec:method_encoders}
\vspace{-1mm}

As shown in Fig.~\ref{fig:model_diagrams}, we consider different encoders which fuse audio and visual information at different stages. %
In all cases, we use the standard transformer architecture~\cite{dosovitskiy_iclr_2021,vaswani_neurips_2017}.%

\vspace{\paravspace}
\paragraph{Early fusion}
We first concatenate the unmasked tokens from the respective modalities, before passing them to a single transformer (Fig.~\ref{fig:model_diagrams}a).
This method is thus an ``early'' fusion~\cite{karpathy_cvpr_2014} of audio and video.
Due to the high masking ratios, $\alpha_a$ and $\alpha_v$ used for audio and video respectively, this is still computationally efficient and allows the encoder to model joint interactions between both modalities.

\vspace{\paravspace}
\paragraph{Separate}
This variant, in contrast, encodes audio and video tokens with two separate encoders each with different parameters.
When using such an encoder, a ``late fusion''~\cite{karpathy_cvpr_2014,simonyan_neurips_2014} of audio and video is performed in the decoder when reconstructing the tokens.
This strategy, however, allows us to obtain separate encoders for each modality after pretraining, which may be advantageous for finetuning on a unimodal (\ie audio-only or video-only) downstream task.

\vspace{\paravspace}
\paragraph{Mid-fusion}
Here, we perform a middle-ground between the previous two approaches (Fig.~\ref{fig:model_diagrams}d).
Denoting the total number of transformer layers as $L$, the first $L - S$ layers are separate for each modality as in the ``Separate'' encoding approach.
The tokens are then concatenated into a single sequence, and then forwarded through a further $S \geq 1$ layers which jointly process both modalities.

\vspace{\paravspace}
\paragraph{Shared}
Finally, we explore coupling the two modalities together via parameter-sharing.
As shown in Fig.~\ref{fig:model_diagrams}c, the unmasked tokens for audio and video are encoded separately, but using the same transformer parameters in both cases.
This is therefore equivalent to the ``Separate'' strategy, where the weights are tied between the encoders.

\vspace{-0.15\baselineskip}
\subsection{Decoders}
\label{sec:method_decoders}
\vspace{-1mm}

In the masked autoencoding framework~\cite{he2022masked, feichtenhofer2022masked, bao2021beit}, the decoder is another transformer that reconstructs the masked tokens given the encoded tokens as context.
The decoder has less capacity than the encoder, to force the encoder to learn discriminative features which can be used for reconstruction. %
Moreover, this  also  improves training efficiency, as mask tokens are also processed by the decoder.

Our decoder can follow any of the encoder structures described previously, whilst also being shallower.
Note that when the ``separate'' encoding strategy is used, the same decoding strategy should not be used as this would amount to two separate, uncoupled masked autoencoder models.

\subsection{Objective} 
\label{sec:method_objective}
\vspace{-1mm}

After encoding the unmasked tokens, and inserting the mask tokens back into the sequence, we obtain the inputs to our decoder, namely $\mathbf{z}_a \in \mathbb{R}^{n_a \times d}$ and $\mathbf{z}_v \in \mathbb{R}^{n_v \times d}$.
Here, $\mathbf{z}_a$ and $\mathbf{z}_v$ correspond to the audio and video modalities, respectively.
Similarly, $n_a$ and $n_v$ are the total number of audio and video tokens respectively.

\vspace{\paravspace}
\paragraph{Joint Reconstruction}
This is a straightforward extension of the masked autoencoding objective~\cite{he2022masked}.
We reconstruct the original audio and video inputs that correspond to the mask tokens in $\mathbf{z}_a$ and $\mathbf{z}_v$ respectively, with the mean square error.
This is denoted for audio as
\vspace{-0.25\baselineskip}
\begin{equation}
\mathcal L_a(\mathbf{\mathbf{z}_a, \tilde{x}}_a) = \frac{1}{\alpha_a n_a} \sum_{i \in \mathcal{M}_a} || \text{Decode}(\mathbf{z}_{a, i}) - \tilde{\mathbf{x}}_{a, i} ||^{2},
\label{eq:reconstruction}
\end{equation}
where $\mathcal{M}_a$ denotes the set of mask token indices, as we only compute the reconstruction loss on these tokens like~\cite{he2022masked,feichtenhofer2022masked,wei2022masked}.
The reconstruction loss for the video modality is defined similarly, and to avoid additional hyperparameters, we use equal loss weights for both audio and video. %

\vspace{\paravspace}
\paragraph{Modality inpainting}
This objective aims to encourage stronger coupling between the two modalities by reconstructing one modality from the other. %
As shown in Fig.~\ref{fig:model_inpainting}, we reconstruct the input video tubelets, $\mathbf{\tilde{x}}_v$ using the encoded audio, $\mathbf{e}_a$ and video mask tokens, $m_v$. 
Similarly, we reconstruct the input audio tokens, $\mathbf{\tilde{x}}_a$, using the encoded video $\mathbf{e}_v$ and audio mask tokens, $m_a$.
Note that in order to use this objective, we require cross-modal encoding of the audio and video tokens, for example by using the ``Early'' or ``Mid-fusion'' encoding methods.
Otherwise, the video tokens will not have the information necessary to reconstruct the audio tokens, and vice versa.

Another consideration is that there is no correspondence between the two sets of tokens (\ie the $i^{th}$ audio token does not correspond to the $i^{th}$ video token, and the respective sequence lengths of the modalities, $n_a$ and $n_v$, are also typically different).
This poses a challenge for reconstruction, as for example, we do not know which video tokens need to be reconstructed given the encoded audio- and mask video-tokens.
To resolve this, we create an ordering in the sequence fed to our decoder, $\mathbf{z}$, by placing all the encoded tokens at the beginning of the sequence, and all of the mask tokens at the end (Fig.~\ref{fig:model_inpainting}), and then only computing the reconstruction error (Eq.~\ref{eq:reconstruction}) on the mask tokens.

Concretely, this means that for reconstructing video, the token sequence is $\mathbf{z}_v = [\mathbf{e}_a || \mathbf{\hat{m}}_v]$, where $[\cdot || \cdot]$ denotes concatenation.
Note that the mask video token, $m_v$, is repeated $n_v - (1 - \alpha_a) \cdot n_a$ times to form, $\mathbf{\hat{m}_v}$ %
 and the reconstruction error is only computed on these tokens once they are decoded.
Similarly, we reconstruct audio tokens by decoding the sequence $\mathbf{z}_a = [\mathbf{e}_v || \mathbf{\hat{m}}_a]$ where there are $n_a - (1 - \alpha_v) \cdot n_v$ mask tokens.

We experimentally evaluate our proposed pretraining architectures and objectives next.

\vspace{-1mm}
\section{Experimental Evaluation}
\vspace{-1mm}

We first describe our experimental setup in Sec.~\ref{sec:exp_setup}, before presenting ablation studies (Sec.~\ref{sec:exp_ablations}) and comparisons to the state-of-the-art in Sec.~\ref{sec:exp_sota}.
For reproducibility, we will release code and trained models on acceptance.

\vspace{-0.2\baselineskip}
\subsection{Experimental setup}
\label{sec:exp_setup}

\paragraph{Datasets}

We select 3 audiovisual datasets, where the audio and video signals are correlated 
and have been shown to be complementary~\cite{nagrani2021attention, jaegle2021perceiver, wang2020makes}:
VGGSound~\cite{chen2020alarge}, AudioSet~\cite{gemmeke2017audio} and Epic Kitchens~\cite{damen2022rescaling}.
We follow standard protocols for these datasets, which we describe below.

\noindent\quad\textit{VGGSound}\cite{chen2020alarge} is an audiovisual benchmark of almost $200 \, 000$ video clips, that are $10$ seconds long and collected from YouTube.
The videos were selected such that the object that is generating the sound is also visible in the video.
Each video is annotated with one of $309$ classes.
After removing the unavailable URLs from the data set, we end up with $170 \, 000$ samples for training and $14 \, 448$ for testing.

\noindent\quad\textit{AudioSet}~\cite{gemmeke2017audio} is the largest audiovisual dataset, consisting of almost $2$ million videos from YouTube which are $10$ seconds long.
There are $527$ classes, and the mean Average Precision (mAP) is the standard metric as the dataset is multilabel.
The dataset is released as a set of URLs, and after accounting for videos that have been removed, we obtain $1.8$ million training videos and $18 \, 589$ validation examples. %
AudioSet is an extremely imbalanced dataset, and numerous works reporting on it have used complex, batch sampling strategies~\cite{gong2021psla, gong2021ast, xu2022masked} or trained on a smaller, but more balanced subset~\cite{nagrani2021attention} of $500 \, 000$ training videos which we denote as AS500K.
For self-supervised pretraining, we use the full, imbalanced AudioSet dataset, which we denote as AS2M, whereas for finetuning we use AS500K.

\noindent\quad\textit{Epic Kitchens-100}~\cite{damen2022rescaling} consists of $67 \, 000$ egocentric videos spanning 100 hours.
We report results following the standard ``action recognition'' protocol.
Here, each video is labelled with a ``verb'' and a ``noun'', and the top-scoring verb and noun pair predicted by the network form an ``action'', and action accuracy is the primary metric.
Following~\cite{arnab2021vivit, yan2022multiview, nagrani2021attention}, we predict both verb and noun classes with a single network with two ``heads'', and equal loss weights. %

In contrast to the other datasets, Epic Kitchens does not use YouTube videos, and as the videos are egocentric, there is a substantial domain shift between them.
Therefore, Epic Kitchens is well-suited for evaluating the transferability of the feature representations learned on the other datasets.

\vspace{\paravspace}
\paragraph{Implementation details}
Our training hyperparameters are based on the public implementations of unimodal MAE models~\cite{he2022masked, tong2022videomae}.
We use a masking ratio of $\alpha_v = 0.9$ for video and $\alpha_a=0.5$ for audio.
For both modalities, we use random token masking as prior works found this to outperform other masking patterns~\cite{he2022masked, feichtenhofer2022masked, xu2022masked}.

Following~\cite{gong2021ast, nagrani2021attention}, audio for all datasets is sampled at 16 kHz and converted to a single channel. %
We extract log mel-spectrograms, using 128 frequency bins computed using a 25ms Hamming window with a hop-length of 10ms.
Therefore the input dimension is $128 \times 100t$ for $t$ seconds of audio.
Since we use 8 seconds of audio as~\cite{nagrani2021attention}, and a spectrogram patch size of $16 \times 16$, we have $n_a = 400$ audio tokens. %
For video, following~\cite{tong2022videomae, feichtenhofer2022masked}, we randomly sample 16 frames, which we tokenise with tubelets of size $16 \times 16 \times 2$.
We implemented our model using the Scenic library~\cite{dehghani2021scenic} and JAX~\cite{jax2018github}, and include exhaustive details in the supplementary and code release.

\vspace{\paravspace}
\paragraph{Network architecture}
We use a standard vision transformer~\cite{dosovitskiy_iclr_2021}, namely the ViT-Base and Large models, following the same configurations as~\cite{dosovitskiy_iclr_2021, devlin_naacl_2019}.
When using the ``mid-fusion'' encoder, we use $S = 2$ shared layers between the modalities, and ablate this choice in the supplementary.

\vspace{\paravspace}
\paragraph{Finetuning}
We first pretrain our network architectures from Sec.~\ref{sec:method}, and then evaluate the learned representations by finetuning on target datasets.
The decoder of the model is only used during pretraining, and is discarded for finetuning as shown in Fig.~\ref{fig:teaser}.

For unimodal finetuning, we simply reuse the pretrained encoder of our model, and feed it a sequence of either audio or video tokens.
For audio, this corresponds to AST~\cite{gong2021ast}.
While for video, this is the unfactorised ViViT model~\cite{arnab2021vivit}.

For audiovisual finetuning, we can reuse the same encoder structure as pretraining for encoding both modalities.
Another alternative is to reuse the current state-of-the-art model, MBT~\cite{nagrani2021attention}, and initialise it with our self-supervised pretrained parameters. %
The MBT model is analogous to our ``Mid-fusion'' approach, it consists of separate transformer encoders for the audio and video modalities, with lateral connections between the two modalities after a predefined number of layers.
When we use the ``Separate'' encoding strategy, we can initialise each stream of MBT with the corresponding encoder from the pretrained model.
And when we use the ``Early fusion'' or ``Shared'' encoding methods, we can initialise each stream of MBT with the same encoder weights which are then untied from each other during training.
In fact, the MBT authors used this method for initialising from ImageNet-21K with supervised pretraining~\cite{nagrani2021attention}.

As detailed in the supplementary, and perhaps surprisingly, we found that audiovisual finetuning with the MBT model performed the best, regardless of the pretraining architecture.
We therefore use this approach in all our audiovisual finetuning experiments.
Morever, as MBT is the current state-of-the-art multimodal fusion model, it allows us to fairly compare our audiovisual self-supervised pretraining to supervised pretraining in state-of-the-art comparisons.

\subsection{Ablation Studies}
\label{sec:exp_ablations}
\vspace{-1mm}

\paragraph{Pretraining architecture}

\begin{table}[t]
\caption{Ablation study of different pretraining architectures.
Models are pretrained from random initialisation for 400 epochs on VGGSound, and then finetuned on the same dataset.
}
\vspace{-0.6\baselineskip}
\centering
\scalebox{0.82}{
	\begin{tabular}{llccc}
	\toprule
	Encoder & Decoder & Audio-only & Video-only & Audiovisual \\
	\midrule
	 
	Early fusion            & Shared     &     $55.5$     &   $46.5$   &  $62.2$         \\
	Shared			  & Shared     &     $55.5$ 	&   $44.7$ 		& $62.5$ \\
	Separate          & Shared     &     $55.4$     &   $48.9$   	&  $63.0$         \\
	Mid-fusion        & Shared     &     $55.8$     &   $48.5$   	&  $63.5$         \\
	\midrule
	Mid-fusion        & Early      &     $55.5$     &   $48.5$   &  $63.3$         \\
	Mid-fusion        & Separate   &     $55.5$     &   $47.4$   &  $63.4$       \\
	Mid-fusion        & Shared     &     $55.8$     &   $48.5$   &  $63.5$         \\	
	\bottomrule
	\end{tabular}
}
\vspace{-\baselineskip}

\label{tab:ablation_pretraining_arch}
\end{table}

Table~\ref{tab:ablation_pretraining_arch} compares the different pretraining architectures, described in Sec.~\ref{sec:method_encoders} and~\ref{sec:method_decoders}.
In this experiment, we pretrained all models from random initialisation on both audio and video on the VGGSound dataset using the ``Joint Reconstruction'' objective (Sec.~\ref{sec:method_objective}) and the ViT-Base backbone for 400 epochs.
We then evaluated the learned representations by finetuning on VGGSound.
As shown in Tab.~\ref{tab:ablation_pretraining_arch}, we performed audio-only, video-only and audiovisual finetuning (where we fuse both modalities) using the same pretrained model in all cases.

We observe that the different encoder architectures perform similarly for audio-only finetuning.
However, there is more of a difference for video-only finetuning, with ``Separate'' and ``Mid-fusion'' performing the best.
For audiovisual finetuning, we also find that the ``Separate'' and ``Mid-fusion'' encoder strategies perform the best, with a sizable difference to the other approaches.
A reason for this is that these encoding strategies closely follow the architecture of MBT~\cite{nagrani2021attention}, the current state-of-the-art multimodal fusion model, which we initialise with our pretrained weights.
Moreover, separate parameters and processing streams for each modality increase model capacity. %

In terms of the decoder architecture, the bottom part of Tab.~\ref{tab:ablation_pretraining_arch} shows that the ``Shared'' decoder strategy outperforms the ``Early'' and ``Separate'' approaches by a small margin.
Parameter sharing between the two modalities (as in the ``Shared'' approach) may be more beneficial in the decoder than in the encoder, as it allows better coupling between audio and video: A shared decoder requires the encoded audio tokens to contain information from the video tokens (and vice versa), in order to be able to reconstruct both sets of tokens using the same parameters.
Based on Tab.~\ref{tab:ablation_pretraining_arch} (and additional AudioSet experiments in the supplementary), we use the ``Mid-fusion'' encoder, and ``Shared'' decoder for future experiments unless otherwise stated.

\vspace{\paravspace}
\paragraph{Objective}

\begin{table}[t]
\vspace{-0.25\baselineskip}
\caption{Ablation study of different pretraining objectives.
Models are pre-trained for 400 epochs on VGGSound with the ``Early fusion'' encoder, and ``Shared'' decoder architecture.
}
\vspace{-0.6\baselineskip}
\scalebox{0.78}{
	\begin{tabular}{lcccc}
	\toprule
	Objective & Audio-only & Video-only & Audiovisual \\
	\midrule 
	Joint reconstruction                    &   $55.5$     &   $46.5$     &  $62.2$         \\ %
	Inpainting (video from audio)             &   $51.5$     &   $39.9$     &  $58.4$ \\  %
	Inpainting (audio from video)             &  $52.5$      & $38.1$       & $58.2$\\ %
	Inpainting (both modalities)            & $54.1$       & $38.6$       & $58.4$\\    %
	\bottomrule
	
	\end{tabular}
}
\vspace{-1.5\baselineskip}
\label{tab:ablation_objective}
\end{table}

Table~\ref{tab:ablation_objective} compares our ``Joint reconstruction'' and ``Modality inpainting'' objectives (Sec.~\ref{sec:method_objective}).
In this experiment, we fix our architecture to the 
``Early fusion''  encoder and ``Shared'' decoder for simplicity. %
For modality inpainting, as we reconstruct audio tokens from the encoded video tokens, and vice versa, we cannot use ``Separate'' encoders -- it is necessary for the encoded audio tokens to contain information about the video tokens and vice versa.
And indeed, training did not converge in this setting. %

Table~\ref{tab:ablation_objective} shows that using the ``Joint reconstruction'' objective outperforms ``Modality inpainting''.
A possible explanation is that ``Modality inpainting'' is more challenging, as the model is tasked with cross-modal reconstruction without using any encoded tokens of the target modality.

Nevertheless, analysis of the inpainting objective revealed interesting insights.
Reconstructing video from audio results in better video representations (as shown by the higher video-only finetuning number).
Similarly, reconstructing audio from video yields better audio-only representations. 
Inpainting both audio from video, and video from audio, was difficult to train, and also required careful learning rate tuning to converge (as illustrated by learning curves in the supplementary).
Inpainting both modalities however yields better audio representations, but slightly worse video representations, than inpainting a single modality.
Note that we use equal weights for the video inpainting and audio inpainting losses in order to reduce the number of pretraining hyperparameters.
Future work remains to further improve the potential of this objective, for example by using a combination of our ``Joint Reconstruction'' and ``Modality inpainting'' losses instead.

As the ``Joint reconstruction'' objective outperforms ``Modality inpainting'', and is also simpler to implement and train, we use it for the remainder of our experiments.

\vspace{\paravspace}
\paragraph{Comparing multimodal and unimodal pretraining}

\begin{table}[t]
	\vspace{-0.25\baselineskip}
\caption{Comparison of single-modality and audiovisual pretraining on VGGSound.
	We use a ViT-Base model with 400 epochs of pretraining.
    AudioMAE and VideoMAE refer to an MAE pretrained only on audio and video respectively.
    } 
	\vspace{-0.6\baselineskip}
\centering
\scalebox{0.9}{
    \begin{tabular}{lccc}
    \toprule
               Pretraining                 & Audio only  & Video only & Audiovisual \\ 
                                \midrule
    		AudioMAE            &  $55.7$    &  $42.1$    &   $58.3$    \\
    		VideoMAE            &  $52.8$    &  $49.3$    &   $62.1$   \\
    		Audiovisual MAE            &  $55.8$    &  $48.5$  &  $63.5$   \\ \bottomrule

    \end{tabular}
}
\vspace{-0.8\baselineskip}
\label{tab:ablation_unimodal_vs_multimodal}
\end{table}

Table~\ref{tab:ablation_unimodal_vs_multimodal} compares our Audiovisual MAE to pretraining on audio only (AudioMAE), and video only (VideoMAE). 

When finetuning with audio-only or video-only, our Audiovisual MAE performs similarly to AudioMAE and VideoMAE, respectively.
However, we observe substantial benefits for audiovisual finetuning tasks where we improve by 5.2 and 1.4 points compared to AudioMAE and VideoMAE respectively.
Note that when performing audiovisual finetuning with a single-modality MAE we initialise both modality-streams of MBT~\cite{nagrani2021attention} with the parameters of either AudioMAE or VideoMAE, analogously to how the original authors initialised from ImageNet-pretrained models.

Therefore, we conclude that audiovisual pretraining is especially beneficial for audiovisual finetuning, and still effective if one is interested in unimodal downstream tasks.

\vspace{\paravspace}
\paragraph{Transferability of learned representations}

Our experiments thus far have pretrained and finetuned on the same dataset.
To evaluate the transferability of learned representations, we pretrain and finetune across different datasets as shown in Tab.~\ref{tab:ablation_transfer}. %
In this experiment, we pretrain for an equivalent number of iterations on both datasets.
This is 800 epochs on VGGSound, and 80 epochs on AudioSet as the dataset is 10 times larger than VGGSound.

\begin{table}[t]
\caption{Transferability of learned representations by pretraining and finetuning across different datasets.
    We pretrain for the equivalent number of iterations on both datasets using ViT-Large as the backbone.
    Following standard protocol, we report Top-1 accuracy for VGGSound, mean Average Precision for AudioSet, and Top-1 action accuracy for Epic Kitchens.
	}
	\vspace{-0.6\baselineskip}
	\scalebox{0.87}{
		\begin{tabular}{lccc}
		\toprule
			\diagbox{Pretrain}{Finetune} 
			& VGGSound & AudioSet & Epic Kitchens \\ \midrule
			VGGSound         &    \textbf{65.0}      &   51.2       &    \textbf{45.5}         \\
			AudioSet           &     64.7      &    \textbf{51.3}       &     43.5        \\
			\bottomrule
		\end{tabular}
	}
	\vspace{-0.5\baselineskip}
\label{tab:ablation_transfer}
\end{table}

When evaluating on either VGGSound or AudioSet, we observe small differences between pretraining on either dataset, indicating that the learned representations transfer across both datasets.
As expected, pretraining and finetuning on the same dataset still produces the best results.

However, we observe larger differences on Epic Kitchens, where pretraining on VGGSound improves action accuracy by 2\% compared to pretraining on AudioSet.
Note that the domain of Epic Kitchens, which consists of videos taken by egocentric cameras in household environments, is quite different to that of the YouTube videos which comprise VGGSound and AudioSet.
Therefore, evaluating transfer performance on Epic Kitchens is a challenging task.

The fact that pretraining on VGGSound performs better than the substantially larger AudioSet dataset suggests that the number of iterations of pretraining are more important than the actual size of the pretraining dataset, in line with some of the observations made by~\cite{tong2022videomae}, and an additional experiment in our supplementary.

\vspace{\paravspace}
\paragraph{Effect of pretraining epochs}
\begin{table}[t]
	\caption{
		Effect of number of pretraining epochs, using ViT-Large.
		We pretrain on VGGSound, and evaluate with audiovisual finetuning on both VGGSound and Epic Kitchens, to evaluate the transferability of the learned representation.
	}
	\vspace{-0.5\baselineskip}
	\centering
	\begin{tabular}{lcccc}
	\toprule
		Epochs           &   200    &  400   & 800              & 1200\\ \midrule 
		VGGSound         &   63.2   &  63.9  & \textbf{65.0}    &      64.9  \\
		Epic Kitchens 	 &  41.8    &  42.5  & 45.5             &  \textbf{46.0} \\
	\bottomrule
	\end{tabular}
	\label{tab:ablation_epochs_effect}
	\vspace{-\baselineskip}
\end{table}

Table~\ref{tab:ablation_epochs_effect} compares the effect of the number of pretraining epochs on downstream, audiovisual finetuning accuracy.
When pretraining and finetuning on the same VGGSound dataset, we observe that accuracy saturates at 800 pretraining epochs.
However, when finetuning on the Epic Kitchens dataset, we can see that pretraining for longer, up to 1200 epochs, is still beneficial and we do not yet see saturation.
The consistent improvements from pretraining for longer are in line with previous works on unimodal masked autoencoders~\cite{he2022masked,tong2022videomae,feichtenhofer2022masked}.

\vspace{\paravspace}
\paragraph{Scaling up the backbone}

Table~\ref{tab:ablation_model_size} compares the ViT-Base and ViT-Large backbones for audiovisual finetuning on VGGSound.
We consider two scenarios: First, pretraining with our proposed Audiovisual MAE on VGGSound, and second, using supervised pretraining with ImageNet-21K as done by the current state-of-the-art, MBT~\cite{nagrani2021attention}.

\begin{table}[t]
	\vspace{-0.1\baselineskip}
	\caption{
		Effect of scaling up the backbone architecture from ViT-Base to ViT-Large, for both supervised and self-supervised pretraining.
		We report audiovisual accuracy on VGGSound.
		Supervised pretraining initialisation results are obtained from MBT~\cite{nagrani2021attention}.
	}
	\vspace{-0.6\baselineskip}
	\centering
	\begin{tabular}{l cc}
		\toprule
		Initialisation & Base & Large \\
		\midrule
		ImageNet-21K~\cite{nagrani2021attention} (Supervised)    & 64.1   & 61.4     \\
		Audiovisual MAE (Self-supervised)  & \textbf{64.2}   & \textbf{65.0}    \\
		\bottomrule
	\end{tabular}
	\label{tab:ablation_model_size}
	\vspace{-1.25\baselineskip}
\end{table}

With our audiovisual pretraining, we observe a solid improvement from scaling up our model backbone from ViT-Base to ViT-Large. %
MBT~\cite{nagrani2021attention}	which uses supervised pretraining does not benefit from increasing the model backbone to ViT-Large, and in fact, experiences heavy overfitting which reduces its performance.
Note that our improved accuracy is not due to additional regularisation during finetuning to counter overfitting, since we followed the public MBT code and used the same regularisers (stochastic depth~\cite{huang_stochasticdepth_eccv_2016}, mixup~\cite{zhang_mixup_iclr_2018}, label smoothing~\cite{szegedy_cvpr_2016}) as detailed in the supplementary.
Our finding that masked pretraining produces more generalisable representations for finetuning larger models is also in line with~\cite{he2022masked}.

\vspace{\paravspace}
\paragraph{Additional baseline}

We consider another baseline, where we train two separate MAE models, one audio-only and the other video-only on VGGSound for 800 epochs, and use this to initialise an audiovisual MBT model which we finetune on VGGSound.
As detailed in the supplementary, this baseline obtains an audiovisual accuracy of 63.3 on VGGSound, compared to our Audiovisual MAE model which achieves 64.2, thereby showing the benefits of joint pretraining of both audio and video.

\vspace{\paravspace}
\paragraph{Qualitative Results} We include visualisations of our reconstructions from pretraining in the supplementary.

\subsection{State-of-the-art comparisons}
\label{sec:exp_sota}
\vspace{-1mm}

	\begin{table*}[tb]
	\vspace{-0.75\baselineskip}
	\caption{
		State-of-the-art comparisons on audiovisual datasets.
		Previous methods use supervised pretraining on additional data including ImageNet-21K (Im21K), ImageNet-1K (Im1K), AudioSet (AS) and Kinetics 400 (K400).
		Our approach, in contrast, is self-supervised, and uses no labelled data beyond the target dataset. 
		We denote audio-only, video-only and audiovisual finetuning as ``A'', ``V'' and ``AV'' respectively.
		The top-scoring entry is in bold, whilst the second-highest is underlined.
		\vspace{-0.8\baselineskip}
	}
	\begin{subtable}{0.49\linewidth}
		\caption{
			VGGSound. We report Top-1 accuracy.
		}
		\centering
		\vspace{-0.4\baselineskip}
		\scalebox{0.93}{
			\begin{tabular}{llccc}
				\toprule
				Method          & Pretraining & A & V & AV \\ \midrule 
				Kazakos~\etal~\cite{kazakos2021slow} & Sup. Im1K & 52.5 & -- & --\\
				PlayItBack~\cite{stergiou2022play} & Sup. Im21K & 53.7 & -- & --\\
				PolyViT~\cite{likhosherstov2021polyvit} & Sup. Im21K, AS & \underline{55.1} & -- & -- \\  %
				MBT~\cite{nagrani2021attention}       & Sup. Im21K  &   52.3   &  \textbf{51.2}      &   \underline{64.1}          \\
				\midrule
				Ours & SSL VGGSound & \textbf{57.2} & \underline{50.3} & \textbf{65.0} \\  %
				\bottomrule
			\end{tabular}
		}
		\label{tab:sota_vggsound}
	\end{subtable} \hfill
	\begin{subtable}{0.49\linewidth}
		\caption{
			AudioSet.
			We report the mAP for audiovisual fusion models.
		}
		\vspace{-0.4\baselineskip}
		\centering
		\scalebox{0.83}{
			\begin{tabular}{lllccc}
				\toprule
				Method          & Pretraining & Training set & A & V & AV \\ \midrule 
				GBlend~\cite{wang2020makes} 	 & Im1K & AS-2M  	 	 & 32.4 & 18.8 & 41.8 \\  %
				Perceiver~\cite{jaegle2021perceiver} & None & AS-2M  		& 38.4 & 25.8 & 44.2 \\  %
				PerceiverIO~\cite{jaegle2022perceiver}  & None & AS-2M 	 & -- & -- & 44.9 \\
				Fayek~\etal~\cite{fayek2020large}    & Im1K    & AS-2M		& 38.4 & 25.7 & 46.2 \\
				MBT~\cite{nagrani2021attention}      & Im21K & AS-500K   & \underline{41.5}   &  \textbf{31.3}      &   \underline{49.6}        \\
				\midrule    
    
				Ours  & SSL AS-2M & AS-500K  & \textbf{46.6} & \underline{31.1} & \textbf{51.8} \\
                
				\bottomrule
			\end{tabular}
		}
		\label{tab:sota_audioset}
	\end{subtable}
	\begin{subtable}{0.99\linewidth}
		\vspace{0.25\baselineskip}
		\caption{Epic Kitchens. We report Top-1 accuracies for verbs, nouns and actions (pairs of verbs and nouns).}
		\vspace{-0.4\baselineskip}
		\scalebox{0.94}{
			\begin{tabular}{ll ccc ccc ccc}
				\toprule
				& & \multicolumn{3}{c}{Audio} & \multicolumn{3}{c}{Video} & \multicolumn{3}{c}{Audiovisual} \\ \cmidrule(r){3-5} \cmidrule(lr){6-8} \cmidrule(l){9-11}
				Method & Pretraining & Verb   & Noun   & Action  & Verb   & Noun   & Action  & Verb     & Noun     & Action    \\ \midrule
				Damen~\etal~\cite{damen2022rescaling} & Sup. Im1K & 42.6 & 22.4 & 14.5 & -- & -- & -- & -- & -- & -- \\
				Kazakos~\etal~\cite{kazakos2021slow} & Sup. VGGSound  & 46.1 & 23.0 & 15.2 & -- & -- & -- & -- & -- & -- \\
				PlayItBack~\cite{stergiou2022play} & Sup. Im21K & \underline{47.0} & \underline{23.1} & \underline{15.9} & -- & -- & -- & -- & -- & -- \\  %
				TSM~\cite{lin2019tsm} &  Sup. Im1K + K400 & -- & -- & -- & \underline{67.9} & 49.0 & 38.3 & -- & -- & --\\
				ViViT-L Fact. Encoder ~\cite{arnab2021vivit} & Sup. Im21K + K400 & -- & -- & -- & 66.4 & 56.8 & 44.0 & -- & -- & -- \\
				MotionFormer~\cite{patrick2021keeping} & Sup. Im21K + K400 & -- & -- & -- & 67.0 & \underline{58.5} & 44.5 & -- & -- & -- \\
				MTV~\cite{yan2022multiview} & Sup. Im21K + K400 & -- & -- & -- & 67.8 & \textbf{60.5} & \textbf{46.7} & -- & -- & -- \\
				MBT~\cite{nagrani2021attention} & Sup. Im21K &  44.3 & 22.4 & 13.0 & 62.0 & 56.4 & 40.7 & \underline{64.8} & \textbf{58.0} & \underline{43.4} \\ 
				\midrule
				Ours & SSL VGGSound & \textbf{52.7} & \textbf{27.2} & \textbf{19.7} & \textbf{70.8} & 55.9 & \underline{45.8} & \textbf{71.4} & \underline{56.4} & \textbf{46.0} \\ 
				\bottomrule
			\end{tabular}
		}
		\vspace{-1\baselineskip}
		\label{tab:sota_epic_kitchens}
	\end{subtable}
\end{table*}

We now compare our best models, using a ViT-Large backbone, on the audiovisual VGGSound, Epic Kitchens and AudioSet datasets.
Note that these are only system-level comparisons as most prior, published works use supervised pretraining.
In particular for unimodal finetuning, our architecture is a vanilla ViT model without the modality-specific designs used in other works.
We are nevertheless able to achieve results surpassing, or competitive with, the state-of-the-art on a number of domains and modalities, showing the promise of our self-supervised pretraining.
We also include additional state-of-the-art results on audiovisual event localisation~\cite{tian2018audio} in the supplementary.

\vspace{\paravspace}
\paragraph{VGGSound}

Table~\ref{tab:sota_vggsound} shows that we outperform published methods on the VGGSound dataset. 
Prior %
works on this dataset use supervised pretraining on ImageNet-21K. %
In contrast, we use self-supervised pretraining for 800 epochs, and do not use any additional labelled data.
In the audio-only setting, we improve by 2.1 points over PolyViT~\cite{likhosherstov2021polyvit} which was also trained on AudioSet, and by 4.9 points over MBT. 
In addition, we also achieve 65.0\% Top-1 accuracy on audiovisual finetuning, improving upon~\cite{nagrani2021attention} by 0.9 points.

\vspace{\paravspace}
\paragraph{Epic Kitchens}

We now transfer our VGGSound pretrained model from the previous experiment to the Epic Kitchens dataset.
Epic Kitchens consists of egocentric videos, and hence it presents a challenging domain shift compared to the YouTube videos in our pretraining dataset.
Nevertheless, Tab.~\ref{tab:sota_epic_kitchens} shows that we outperform MBT substantially by 2.6 points on audiovisual finetuning.
On audio, we substantially outperform recent work~\cite{stergiou2022play} by 3.8 points.
This is likely because our self-supervised pretraining provides better initialisation than the ImageNet-pretrained weights typically used by the audio community.

For video-only finetuning, our architecture corresponds to an unfactorised ViViT~\cite{arnab2021vivit}. %
We, however, still outperform the ViViT Factorised Encoder~\cite{arnab2021vivit} which the authors showed was more accurate than the unfactorised model.
The key difference is our initialisation -- our model is initialised with self-supervised pretraining, whereas the other video models in Tab.~\ref{tab:sota_epic_kitchens} used supervised pretraining, first on ImageNet-21K and then Kinetics 400~\cite{arnab2021vivit,yan2022multiview,patrick2021keeping}.
And it is this self-supervised initialisation which enables better generalisation of our model to Epic Kitchens.
Only recent work~\cite{yan2022multiview}, %
outperforms our method for video-only finetuning with a specialised, multi-view architecture and additional supervised pretraining (ImageNet-21K and Kinetics 400).

Note that previous transformer models on this dataset~\cite{arnab2021vivit,yan2022multiview,patrick2021keeping,nagrani2021attention}, are pretrained on ImageNet-21K and have a bias towards ``noun'' classes, performing well on these, and poorly on ``verb'' classes, compared to previous CNN models~\cite{lin2019tsm}.
Intuitively, this is because the ImageNet dataset is labelled with object classes corresponding to nouns, and models finetuned from this initialisation perform well on nouns, but struggle on verbs.
Our self-supervised pretraining, in contrast, does not utilise class labels, and performs significantly better on verb classes across all modalities.

\vspace{\paravspace}
\paragraph{AudioSet}

Table~\ref{tab:sota_audioset} compares our model to other published %
audiovisual models on AudioSet.
We performed self-supervised pretraining on the full AudioSet-2M (AS2M).
Following MBT~\cite{nagrani2021attention}, we finetuned on the AS500K training subset, which is slightly more balanced than the full AS2M.
Other methods which finetune on the full AS2M have used complex minibatch sampling techniques~\cite{wang2020makes}, which we can avoid due to our use of AS500K.
We improve substantially upon previous %
methods on audiovisual finetuning by 2.2 points, and audio-only finetuning by 5.1 points.
Note that we have reported the recently corrected results of~\cite{nagrani2021attention}.

\section{Conclusion and Future Work}
\vspace{-1mm}

We have proposed Audiovisual MAE,
a simple and effective self-supervised approach for learning powerful and generalisable audiovisual representations.
The efficacy of our approach is demonstrated by our state-of-the-art results, in both audiovisual and unimodal downstream tasks, on the VGGSound, AudioSet and Epic Kitchens datasets.

In future work, we aim to leverage more powerful backbone architectures than a standard vision transformer~\cite{dosovitskiy_iclr_2021} and improve our modality inpainting objective, for example, by combining it with the ``Joint reconstruction'' loss term.

\noindent \paragraph{Acknowledgements} We would like to thank David Ross, Xuehan Xiong, Dan Ellis and Aren Jansen for helpful feedback and discussions.

{\small
\bibliographystyle{ieee_fullname}
\bibliography{bibliography}
}

\clearpage

\section*{Appendix}

\appendix

In this appendix, we include additional experimental details and results.
We include additional ablation studies and evaluation in Sec.~\ref{sec:supp_ablations}, details about our experimental hyperparameters in Sec.~\ref{sec:supp_exp_details} and qualitative visualisations in Sec.~\ref{sec:supp_visualisation}.

\section{Additional Experiments and Ablation Studies}
\label{sec:supp_ablations}

This section presents additional experiments and ablation studies and evaluations of our model.
Unless otherwise stated, the experiments are performed using a ViT-Base backbone, pretrained for 400 epochs on VGGSound, using the ``Separate'' encoding and ``Shared'' decoding strategies.

\subsection{Audiovisual event localisation}
In Sec.~4.3 of the main paper, using our learned representations we obtain state-of-the-art results on three downstream classification tasks. 
To show the capabilities of our audiovisual representations in a different downstream task, in Tab.~\ref{tab:rebuttal_supervised_event_localisation} we evaluate on the ``Supervised Event Localisation'' task proposed by \cite{tian2018audio} using a ViT-Base backbone.

We consider two models, one pretrained on VGGSound for 800 epochs, and another pretraind on AudioSet for 80 epochs.
These two models are pretrained for approximately the same number of iterations as AudioSet is about 10 times larger than VGGSound.

To our knowledge, we outperform the best method (concurrent work) on this task, when pretraining on either VGGSound or AudioSet.
We observe that pretraining on VGGSound learns better audiovisual representations overall for this dataset.

\begin{table}[t]
	\caption{Supervised event localisation accuracy on the AVE dataset~\cite{tian2018audio}. 
		We outperform prior work when using a ViViT-Base model, and pretraining on either VGGSound or AudioSet for the equivalent number of itrations (since AudioSet is approximately 10 times larger than VGGSound.
	} 
	\vspace{-0.8\baselineskip}
	\centering
	\scalebox{0.75}{
		\begin{tabular}{lccc}
			\toprule
			& Audio-only  & Video-only & Audiovisual \\ 
			\midrule
			Senocak~\etal \cite{senocak2023event}            &  79.1    &  76.1    &   87.8    \\
			Ours (Audioset, 80 epochs) & \textbf{82.3} & 77.6 & 88.6\\
			Ours (VGGSound, 800 epochs) & 81.3 & \textbf{78.2} & \textbf{90.2} \\
			\bottomrule
		\end{tabular}
	}
	\label{tab:rebuttal_supervised_event_localisation}
\end{table}

\subsection{Pretraining methods for MBT}
\begin{table}[t]
	\centering
	\caption{Comparison of pretraining methods according to model size. Our self-supervised pretraining scales with the model size, unlike supervised pretraining on ImageNet-21K, as used by MBT~\cite{nagrani2021attention}. We report audiovisual finetuning accuracy for VGGsound and mAP for AudioSet.}
	\renewcommand{\arraystretch}{0.8}
	\scalebox{0.8}{
		\begin{tabular}{clcc}
			\toprule
			Model size & Pretraining           & VGGSound & AudioSet \\
			\midrule
			Base	 & Scratch                &     51.0     &  39.9         \\ 
			($172 \times 10^6$  & Supervised, ImageNet-21K 			& 64.1          &   49.6       \\
			params)	 & Self-supervised, ours    			& \textbf{64.2}         &    \textbf{50.0}     \\
			\midrule
			Large		& Scratch                  & 41.6         & 21.5      \\ 
			($611 \times 10^6$	& Supervised, ImageNet-21K 			& 61.4		&  48.2      \\
			params)		& Self-supervised, ours    &  \textbf{65.0}        &  \textbf{51.8}       \\
			\bottomrule
		\end{tabular}
	}
	\vspace{-0.8\baselineskip}
	\label{tab:rebuttal_pretraining_model_size}
\end{table}

Our main contribution is an audiovisual, self-supervised pretraining method.
To show the benefit of our pretraining, for downstream finetuning we used the same model architecture as the current SOTA (MBT~\cite{nagrani2021attention}).

Table~\ref{tab:rebuttal_pretraining_model_size} shows audiovisual recognition performance when training MBT on the target datasets VGGSound and AudioSet for three different pretraining strategies: (1) from scratch (i.e., no pretraining), (2) initializing MBT from a ViT pretrained with supervised image-classification labels on ImageNet-21K (as done in~\cite{nagrani2021attention}), (3) using our proposed mask-based self-supervised pretraining on each target dataset.

Our proposed pretraining, using \textit{only} the target datasets without labels, outperforms the original MBT setup. Notice that the original MBT setup~\cite{nagrani2021attention} is based on pretraining on an \textit{external} dataset different from the target ones, and using expensive labels for millions of examples.
If we train MBT using data from only VGGSound / Audioset, as our method, we must train it from scratch, and the results are significantly worse. 

Table~\ref{tab:rebuttal_pretraining_model_size} also shows that our proposed pretraining scales better with model size than traditional supervised pretraining, in line with results reported in the original MAE paper \cite{he2022masked} on ImageNet.

\subsection{Additional baseline}

\begin{table}[t]
	\caption{
	Additional baseline for Audiovisual MAE on VGGSound. We report the audiovisual finetuning accuracy.
	Note that joint modelling and pretraining by our proposed Audiovisual MAE model outperforms the baseline of pretraining two separate, unimodal MAE models.
	}
	\begin{tabular}{lc}
	\toprule
	Method & AV accuracy \\
	\midrule
	 Separate AudioMAE and VideoMAE & 63.3 \\  %
	 Audiovisual MAE & 64.2 \\ %
	\bottomrule
	\end{tabular}
	\label{tab:additional_baseline}
\end{table}

Table~\ref{tab:additional_baseline} reports an additional baseline for our proposed Audiovisual MAE model.

Here, we train two separate MAE models on audio-only and video-only on VGGSound for 800 epochs, and use this to initialise an MBT model which we then finetune on VGGSound.
This corresponds to a ``Separate'' encoding and decoding strategy, and thus two separate MAEs pretrained in parallel.
We compare this to our proposed Audiovisual MAE model.

As shown in Tab.~\ref{tab:additional_baseline}, our Audiovisual MAE outperforms this baseline, showing the benefits of joint modelling of both audio and video.

\subsection{Masking ratio}

Tables~\ref{tab:ablation_study_mask_ratio_joint},~\ref{tab:ablation_study_mask_ratio_audio} and~\ref{tab:ablation_study_mask_ratio_video} ablate the effect of the masking ratio in the case of audiovisual, audio-only and video-only pretraining respectively.

In all cases, we pretrain for 400 epochs with ViT-Base on VGGSound.
We use the ``Separate'' encoding and ``Shared'' decoding architecture and the ``Joint Reconstruction'' objective.

We observe that the optimal masking ratios for unimodal and multimodal pretraining are correlated.
However, the best masking ratio for video-only for example is 0.95 (Tab.~\ref{tab:ablation_study_mask_ratio_video}), but this is not the best value for audiovisual pretraining as shown in Tab.~\ref{tab:ablation_study_mask_ratio_joint}.

\begin{table}[t] 
	\caption{Ablation study of different mask ratios.
		 	       We use a ViT-Base backbone, ``Separate'' encoding and ``Shared'' decoding, architecture pretrained for 400 epochs with the ``Joint Reconstruction'' objective.
	 	       		The table shows audiovisual finetuning accuracy on VGGSound.
 	       			}
	\vspace{-0.6\baselineskip}
	\centering
		\begin{tabular}{llccc}
			\toprule 
			\diagbox{Video}{Audio}  &  0.3  & 0.5 & 0.7 & 0.8\\
			\midrule
			
			0.7          &  62.4 & 63.4 &  62.2 & 61.6     \\
			0.9          &  63.3 & 63.0 &  63.5 & 62.3   \\
			0.95         &  63.0 & 63.0 &  63.0 & 62.8 \\
			
			\bottomrule
		\end{tabular}
		\label{tab:ablation_study_mask_ratio_joint}
\end{table} 

\begin{table}[t] 
	\caption{Ablation study of mask ratios when pretraining and finetuning on audio-only on VGGSound. 
	}
	\vspace{-0.6\baselineskip}
	\centering
		\begin{tabular}{cc}
			\toprule 
			Mask ratio for audio &  Accuracy \\
			\midrule
			
			0.3       & 55.1  \\   %
			0.5       & 55.7   \\  
			0.7       & 55.5 \\  %
			0.8       & 55.3 \\   %
			
			\bottomrule
		\end{tabular}
		\label{tab:ablation_study_mask_ratio_audio}	
\end{table} 

\begin{table}[tb] 
	\caption{Ablation study of mask ratios when pretraining and finetuning on video-only on VGGSound. 
			 }
	\vspace{-0.6\baselineskip}
	\centering
		\begin{tabular}{cc}
			\toprule 
			Mask ratio for video &  Accuracy \\
			\midrule
			0.7          & 49.1  \\ %
			0.9          & 49.3   \\ %
			0.95         & 49.5 \\ %
			\bottomrule
		\end{tabular}
		\label{tab:ablation_study_mask_ratio_video}
\end{table} 
\begin{figure*}[t]
    \centering
    \vspace{-\baselineskip}
    \includegraphics[width=0.9\linewidth]{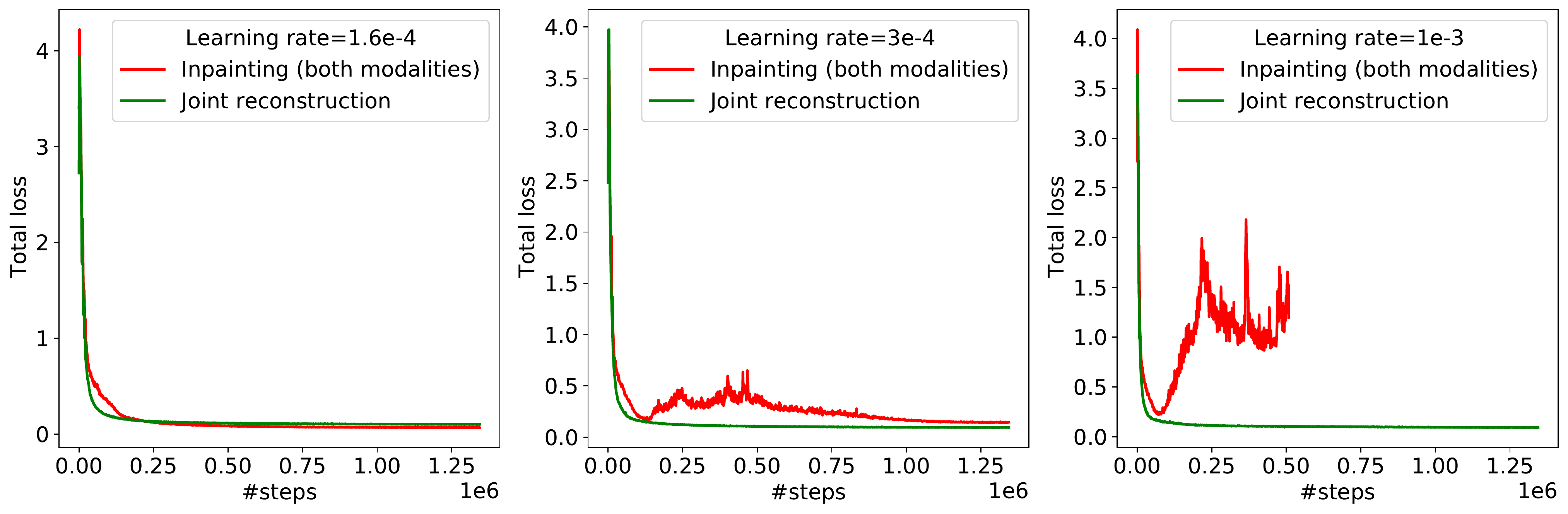} 
    \vspace{-0.7\baselineskip}
    \caption{Learning curves for the ``Joint Reconstruction'' and ``Modality Inpainting'' objectives. Observe how ``Joint Reconstruction'' is stable across a wide range of learning rates. ``Modality Inpainting'', on the other hand, only performs well for a learning rate of $1.6 \times 10^{-4}$, and is unstable at higher values.
    These pretraining experiments were performed on VGGSound for 400 epochs with a batch size of 512.}
	\vspace{-\baselineskip}
    \label{fig:ablation_inpainting_loss_curve} %
\end{figure*}

\subsection{Ablation of audiovisual finetuning architecture}

As mentioned in Sec.~4.1 of the main paper, for audiovisual finetuning, we can either finetune using the original pretraining encoder architecture.
Or, we can instead initialise an MBT~\cite{nagrani2021attention} model.
As shown in Tab.~\ref{tab:ablation_finetuning_arch}, we consistently find that finetuning with an MBT model is better, regardless of the original pretraining architecture.

\subsection{Modality inpainting}

As mentioned in Sec.~4.2 of the main paper, we found that the ``Modality inpainting'' model is difficult to optimise, and requires learning rate tuning in order to train in a stable manner.
This is shown in Fig.~\ref{fig:ablation_inpainting_loss_curve}: The ``Joint reconstruction'' objective is stable across three different learning rate values.
The ``Modality inpainting'' objective, on the other hand, only trains well for one of these learning rates. 
At a higher learning rate of $10^{-3}$, the loss diverges, which is why we stopped training.

\begin{table}[t] 
	\caption{Effect of the finetuning architecture.
		For audiovisual finetuning, we can either finetune using the original encoder architecture, or we can initialise an MBT~\cite{nagrani2021attention} model instead.
		We consistently find that finetuning with an MBT architecture is better, regardless of the original pretraining architecture.
	}
	\vspace{-0.6\baselineskip}
	\centering
		\begin{tabular}{cccc}
			\toprule
			\multicolumn{2}{c}{Pretraining} & \multicolumn{2}{c}{Finetuning} \\
			\cmidrule(r){1-2} \cmidrule(l){3-4}
			Encoder & Decoder &  Pretraining encoder &  MBT\\
			\midrule
			
			Early fusion           & Shared     &    59.4  & 62.2         \\
			Early fusion           & Separate   &    58.1  & 61.1      \\
			Separate          & Shared     &       60.4    & 63.0  \\
			Shared            & Separate   &       58.7    & 61.3 \\
			\bottomrule
		\end{tabular}
		\label{tab:ablation_finetuning_arch}
\end{table}

\subsection{Mid-fusion layer hyperparameter}

\begin{table}[t]  
	\caption{Ablation of $S$, the hyperparameter denoting the number of shared layers when using the ``Mid-fusion'' encoding strategy.
		The experiment is performed on ViT-Base, where there are a total of 12 layers.
		We report audiovisual finetuning accuracy on VGGSound.}
	\centering 
		\begin{tabular}{cc}
			\toprule
			$S$ & Accuracy\\ \midrule 
			$S=1$   &  63.4\\
			$S=2$   &  63.5 \\
			$S=3$   &  63.2 \\  
			$S=4$  &  63.1 \\   
			
			\bottomrule
		\end{tabular}
	\label{tab:ablation_midfusion}
\end{table}

For our mid-fusion architecture (Sec.~3.2 of the main paper), we have an additional hyperparameter $S$, which denotes the number of shared layers.
Table~\ref{tab:ablation_midfusion} ablates this hyperparameter for a ViT-Base model with a total of 12 layers.
As with the other ablation experiments, it was performed on VGGSound whilst pretraining for 400 epochs.

\subsection{Mid-fusion vs Separate encoders on AudioSet}
In Sec.~4.2 of the main paper, we show that the ``Mid-fusion'' encoding strategy slightly outperforms other encoding strategies on audiovisual classification using VGGSound.
Here we compare the ``Mid-fusion'' strategy vs the ``Separate'' encoders strategy on AudioSet, using our best setup consisting of a ViT-Large backbone pre-trained for 120 epochs.
Results in Tab.~\ref{tab:rebuttal_audioset_encoder_comparison} confirm that ``Mid-fusion'' also exhibits slightly better performance on AudioSet.

As noted in the main paper, ``Early fusion'' uses the same model parameters for all modalities, and thus does not allow modality-specific modelling.
The late fusion provided by ``Separate'' encoders, in contrast, does not allocate many parameters to model interactions between modalities.
``Mid-fusion'' is a middle-ground, featuring both modality-specific parameters, and sufficient layers to model inter-modality relations.
The benefits of mid-fusion have also been observed empirically by MBT~\cite{nagrani2021attention} in a supervised setting.
\begin{table}[t]
	\caption{Encoder architecture comparison on AudioSet. Large backbone pretrained for 120 epochs, using a ``Shared'' decoder.} 
	\vspace{-0.7\baselineskip}
	\centering
	\scalebox{0.9}{
		\begin{tabular}{lccc}
			\toprule
			& A  & V & AV \\ 
			\midrule
			Separate encoders            & 46.5 & 30.3 & 51.4 \\
			Mid-fusion					 & \textbf{46.6} & \textbf{31.1} & \textbf{51.8} \\
			\bottomrule
		\end{tabular}
	}
	\label{tab:rebuttal_audioset_encoder_comparison}
\end{table}

\subsection{Pretraining for the same number of iterations on different subsets of VGGSound}
In Sec.~4.2 of the main paper, we saw that pretraining on VGGSound leads to better performance on Epic Kitchens than pretraining on the substantially larger AudioSet, when using 10x epochs for VGGSound in order to keep the number of training iterations roughly constant.
This suggests that the number of iterations of pretraining are more important than the actual size of the pretraining dataset, in line with
some of the observations made by \cite{tong2022videomae}.

For an additional comparison, in Tab.~\ref{tab:rebuttal_table_epochs}, we conduct a similar experiment now utilising different subsets of VGGSound. 
In particular, we compare pretraining a ViT-Base backbone on the full VGGSound for 400 epochs, with pretraining on half of VGGSound for 800 epochs, thus keeping the number of training iterations constant.
The similar finetuning results of Tab.~\ref{tab:rebuttal_table_epochs} support the hypothesis posed in Sec.~4.2 that the number of pretraining iterations is more critical than the size of the pretraining dataset.
El-Nouby~\etal~\cite{el2021large} and Tong~\etal~\cite{tong2022videomae} have also observed self-supervised pretraining performing well on smaller datasets.
We aim to study exactly how much pretraining data is needed further in future work.
\begin{table}[t]
	\caption{Pretraining for the same number of iterations on different subsets of VGGSound produces similar finetuning results.
	} 
	\vspace{-0.8\baselineskip}
	\centering
	\scalebox{0.9}{
		\begin{tabular}{lccc}
			\toprule
			& A  & V & AV \\ 
			\midrule
			VGGSound-50\% for 800 epochs   & $55.5$  & $48.5$ & $63.4$   \\
			VGGSound-100\% for 400 epochs & $55.8$     &   $48.5$   &  $63.5$   \\
			\bottomrule
		\end{tabular}
	}
	\label{tab:rebuttal_table_epochs}
\end{table}

\subsection{Audio-only linear evaluation of different encoder architectures}

\begin{table}[t]
	\caption{Comparison of different pretraining architectures. We show audio-only downstream evaluation on VGGSound.
	}
	\centering
	\scalebox{0.93}{
		\begin{tabular}{llcc}
			\toprule
			Encoder & Decoder                  & Linear probing & Full finetuning \\
			\midrule
			Early fusion & Shared                   &   26.2              & 55.5                \\
			Shared & Shared                			&   27.6              & 55.5 \\
			Separate & Shared                		&   27.6              & 55.4 \\
			Mid-fusion & Shared						 &  \textbf{27.8} 			  & \textbf{55.8} \\
			\bottomrule
		\end{tabular}
	}
	\label{tab:rebuttal_linear_probing}
\end{table}

In Table 1, we saw that different encoder architectures perform similarly for audio-only finetuning.
We analysed this effect further in Table~\ref{tab:rebuttal_linear_probing} by doing linear probing instead.
``Early-fusion'' performs markedly worse in this case, but the other encoder architectures perform similarly.
This suggests that ``early-fusion'' learns different audio representations, but the effect is concealed by fully finetuning the network.
Mid- and late-fusion seem to learn similar representations though.

\subsection{Computational cost}
\begin{table}[t]
	\caption{
		Pretraining time analysis on VGGSound, using identical hardware.
		We also report audiovisual finetuning accuracy.
	}
		\renewcommand{\arraystretch}{0.8}
		\scalebox{0.78}{
			\begin{tabular}{m{4.5cm} x{1.6cm} x{1.5cm} x{1.5cm}}
				\toprule
				Pretraining & Epochs /    & Total time & AV \\
				& Iterations  & (hours)    & Accuracy \\
				
				\midrule
				AudioMAE                      &  800 / 268K             & 59.0 &    58.3        \\
				VideoMAE                      &  800 / 268K             & 84.4 &    62.1        \\
				Separate Audio \& Video MAEs &  800 / 268K             & 143.4 &   63.3        \\
				\midrule
				Audiovisual MAE (ours)         & 800 / 268K             & 89.2 &    64.2		 \\
				\bottomrule
			\end{tabular}
		}
		\label{tab:rebuttal_pretraining_cost}
	\end{table}

Table~\ref{tab:rebuttal_pretraining_cost} compares the wallclock training time of our proposed Audiovisual MAE to separately training audio-only and/or video-only MAEs.
Audiovisual pretraining is only marginally more expensive than video-only pretraining, and provides substantial accuracy gains.
Moreover, we showed in Table~3 that audiovisual pretraining is just as effective for unimodal downstream tasks.
We also significantly outperform the baseline of training separate audio-only and video-only MAEs.

\section{Experimental Details}
\label{sec:supp_exp_details}

In this section, we provide exhaustive details of our experimental setup.
We will also release pretraining code and models, and also finetuning code and models upon acceptance.
Our models are trained using 32 GPU (Nvidia V100) or Cloud TPU v3 accelerators, using the JAX~\cite{jax2018github} and Scenic~\cite{dehghani2021scenic} libraries.

\subsection{Pretraining hyperparameters}

Table~\ref{tab:hyperparams_pretraining} details our hyperparameters for pretraining Audiovisual MAE models.
Note that we use the same pretraining hyperparameters for different datasets.
And we only vary the number of epochs according to the dataset.
Our hyperparameters are based on those of~\cite{he2022masked, tong2022videomae, feichtenhofer2022masked}.
Note that we linearly scale our learning rate with the batch size~\cite{goyal2017accurate}, and we show the learning rate for the reported batch size.
Additionally, we can use a larger batch size during pretraining due to the high masking ratio for Audiovisual MAE pretraining.
As for data normalization, for RGB frames, we followed ViViT~\cite{arnab2021vivit} and zero-centered inputs, from the interval $[0, 255]$ to $[-1, 1]$.
For audio, we followed MBT~\cite{nagrani2021attention}, and did not normalise the log-mel spectrograms.

Table~\ref{tab:hyperparams_decoder} also lists the configuration of the decoders that we use whilst pretraining. 
These were set following~\cite{he2022masked,feichtenhofer2022masked,tong2022videomae}.

\subsection{Finetuning hyperparameters}

Tables~\ref{tab:hyperparams_vggsound},~\ref{tab:hyperparams_audioset} and~\ref{tab:hyperparams_epic_kitchens} show our finetuning hyperparameters for the VGGSound, AudioSet and Epic Kitchens datasets respectively.
We typically use the same hyperparameters across different datasets.
However,  we found that audio-only finetuning sometimes required greater regularisation (also noted earlier by~\cite{wang2020makes}), which is why we used a higher Mixup coefficient for it. %

\begin{table}[t] 
	\caption{Pretraining hyperparameters}
	\vspace{-0.6\baselineskip}
	\centering
		\begin{tabular}{lc}
			\toprule 
			Configuration          & Value \\
			\midrule 
			Optimizer              &    Adam    \\
			Optimizer momentum     &   $\beta_1, \beta_2=0.9, 0.95$ \\ 
			Weight decay           & 0 \\
			Base learning rate     & $3 \times 10^{-4}$ \\ %
			Learning rate schedule & cosine decay   \\ 
			Warm-up epochs         &  40  \\ 
			Augmentation           & None  \\
			Batch size             & 512    \\ 
		 
			\bottomrule 
		\end{tabular} 
		\label{tab:hyperparams_pretraining}
\end{table}

\begin{table}[t]  
	\caption{Hyperparameters of our decoder used during pretraining. We change the size of our decoder based on the size of the encoder, and use hyperparameters following~\cite{he2022masked, feichtenhofer2022masked, tong2022videomae}}
	\centering
	\vspace{-0.6\baselineskip}
	\begin{tabular}{lcc}
		\toprule
		&  Base  & Large   \\ \midrule 
		Hidden dimension   &  384 & 512 \\
		Number of layers    &  4 & 4 \\
		Number of heads     &  6 & 8 \\   
		MLP dimension       & 1536  & 2048 \\ 
		\bottomrule
	\end{tabular}
	\label{tab:hyperparams_decoder}
\end{table}

\begin{table}[t] 
	\caption{VGGSound finetuning hyperparameters}
	\vspace{-0.6\baselineskip}
	\centering
		\begin{tabular}{lccc}
			\toprule  
			Configuration        & A & V & AV \\
			\midrule
			Number of video frames     & -- & 32 & 32\\
			Spectrogram audio length (seconds)    & 8 & -- & 8  \\
			\midrule
			Optimizer              &  \multicolumn{3}{c}{SGD}  \\
			Optimizer momentum        &  \multicolumn{3}{c}{0.9} \\
			Layerwise decay~\cite{bao2021beit,clark2020electra}    &  \multicolumn{3}{c}{0.75} \\
			Base learning rate          & \multicolumn{3}{c}{0.8} \\
			Learning rate schedule     & \multicolumn{3}{c}{cosine decay}\\
			Gradient clipping        & \multicolumn{3}{c}{1.0}    \\  
			Warm-up epochs          & \multicolumn{3}{c}{2.5} \\
			Epochs                 & \multicolumn{3}{c}{50} \\
			Batch size             & \multicolumn{3}{c}{64} \\
			\midrule 
		    SpecAugment \cite{park2019specaugment}   & \checkmark  & -- & \checkmark \\
			Mixup $\alpha$ \cite{zhang_mixup_iclr_2018}           &   \multicolumn{3}{c}{0.5} \\
			Stochastic depth  \cite{huang_stochasticdepth_eccv_2016}    &  \multicolumn{3}{c}{0.3}   \\
			Label smoothing  \cite{szegedy_cvpr_2016}          &  \multicolumn{3}{c}{0.3}   \\
			\bottomrule
		\end{tabular}
		\label{tab:hyperparams_vggsound}
\end{table}

\begin{table}[th] 
	\caption{Epic Kitchens finetuning hyperparameters}
	\vspace{-0.6\baselineskip}
	\centering
		\begin{tabular}{lccc}
			\toprule 
			Configuration        & A  & V & AV \\
			\midrule
			Number of video frames     & -- & 32 & 32\\
			Spectrogram audio length (seconds)    & 8 & -- & 8  \\
			\midrule
			Optimizer              &  \multicolumn{3}{c}{SGD}   \\
			Optimizer momentum        &   \multicolumn{3}{c}{0.9} \\
			Layerwise decay~\cite{bao2021beit, clark2020electra}    &  \multicolumn{3}{c}{0.75}  \\
			Base learning rate          &  \multicolumn{3}{c}{1.2} \\
			Learning rate schedule     &  \multicolumn{3}{c}{cosine decay} \\
			Gradient clipping        &   \multicolumn{3}{c}{1.0}  \\  
			Warm-up epochs          &   \multicolumn{3}{c}{2.5} \\
			Epochs                 &   \multicolumn{3}{c}{50} \\
			Batch size             &   \multicolumn{3}{c}{64} \\
			\midrule 
			Random time shifting     & \checkmark     & --    &  \checkmark  \\
		    SpecAugment \cite{park2019specaugment}   & \checkmark  & -- & \checkmark \\
			Mixup $\alpha$ \cite{zhang_mixup_iclr_2018}           & 1.25 & 0.5 & 0.5 \\
			Stochastic depth  \cite{huang_stochasticdepth_eccv_2016}    &  \multicolumn{3}{c}{0.3} \\
			Label smoothing  \cite{szegedy_cvpr_2016}          &  \multicolumn{3}{c}{0.3} \\
			\bottomrule
		\end{tabular}
		\label{tab:hyperparams_epic_kitchens}
\end{table}

\begin{table}[th] 
	\caption{ AudioSet finetuning hyperparameters
	}
	\vspace{-0.6\baselineskip}
	\centering
		\begin{tabular}{lccc}
			\toprule
			Configuration        & A &  V & AV \\
			\midrule
			Number of video frames     & -- & 32 & 32\\
			Spectrogram audio length (seconds)    & 10 & -- & 10  \\
			\midrule
			Optimizer              & \multicolumn{3}{c}{SGD}         \\
			Optimizer momentum        & \multicolumn{3}{c}{0.9}         \\
			Layerwise decay~\cite{bao2021beit, clark2020electra}    & \multicolumn{3}{c}{0.75}      \\
			Base learning rate          & \multicolumn{3}{c}{1.6}  \\
			Learning rate schedule     & \multicolumn{3}{c}{cosine decay} \\
			Gradient clipping        & \multicolumn{3}{c}{1.0} \\  
			Warm-up epochs          & \multicolumn{3}{c}{2.5} \\
			Epochs                 & \multicolumn{3}{c}{50} \\
			Batch size             & \multicolumn{3}{c}{128} \\
			\midrule
			Random time shifting     & \checkmark     & -    &  \checkmark  \\
			SpecAugment \cite{park2019specaugment}   & \checkmark     & -  & \checkmark \\
			Mixup $\alpha$ \cite{zhang_mixup_iclr_2018}           & 1.25  & 0.5 & 0.5 \\ %
			Stochastic depth  \cite{huang_stochasticdepth_eccv_2016}    & \multicolumn{3}{c}{0.3} \\
			Label smoothing  \cite{szegedy_cvpr_2016}      & \multicolumn{3}{c}{0.3} \\
			\bottomrule
		\end{tabular}
		\label{tab:hyperparams_audioset}
\end{table}

For audio, we use two modality-specific regularisers.
Firstly, we apply SpecAugment~\cite{park2019specaugment} following the settings used in previous works~\cite{nagrani2021attention, gong2021ast}.
We also apply random time shifting on the spectrogram, which involves circularly shifting the audio spectrogram by a time offset sampled from a uniform distribution.
As mentioned in Sec.~4.2 of the main paper, we are not adopting any dataset balancing techniques for AudioSet.
Instead, we finetuned on the AS500K training subset, which is slightly more balanced than the full AS2M (and also smaller, hence faster to process).
We also use a larger batch size for AudioSet since it is a larger dataset.

Note that prior work that we compare to, such as MBT~\cite{nagrani2021attention}, used the same regularisers as we do (stochastic depth, mixup, label smoothing).
Also following standard practice~\cite{arnab2021vivit,carreira_cvpr_2017,nagrani2021attention}, we process multiple views of the input video, averaging the results of 4 views for every evaluation example.

\section{Qualitative Results}
\label{sec:supp_visualisation}

\begin{figure*}[t]
    \centering
    \vspace{-0.5\baselineskip}
    \includegraphics[width=0.95\linewidth]{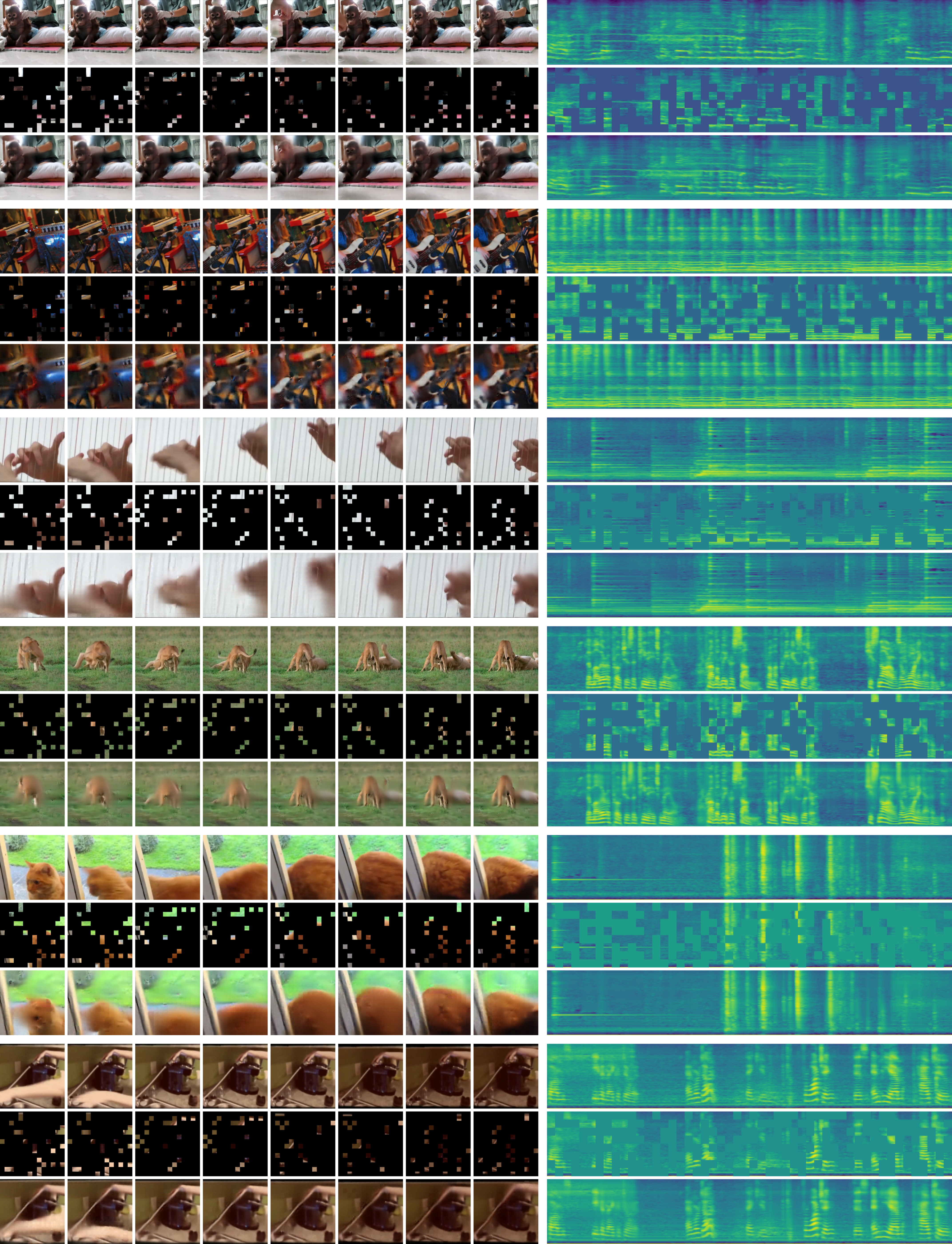} 
    \vspace{-0.7\baselineskip}
    \caption{ 
    	Examples of reconstructions of our model, trained with the ``Joint reconstruction'' objective on AudioSet.
    	We show video frames on the left, and audio spectrograms on the right.
    	The first row shows the original input, the second the input after masking, and the final row shows the reconstruction produced by the model. For the unmasked patches in the reconstruction, we show the original input.
    	Note that the model is pretrained with 16 video frames, and we show 8 here for clarity.
    	This figure is best viewed on screen, zoomed in.
	}
	\vspace{-\baselineskip}
    \label{fig:qualitative_examples}
\end{figure*}

Figure~\ref{fig:qualitative_examples} shows examples of reconstructions of our model trained with the ``Joint reconstruction'' objective on the AudioSet dataset.

\end{document}